\pdfoutput=1
\documentclass[letterpaper,10pt,conference]{ieeeconf}
\IEEEoverridecommandlockouts
\usepackage[utf8]{inputenc}
\usepackage[T1]{fontenc}
\usepackage{tgtermes}
\usepackage{tgcursor}
\usepackage{microtype}
\usepackage{graphicx}
\usepackage{amsmath,amssymb}
\usepackage{booktabs,tabularx,array,multirow}
\usepackage{cite}

\usepackage[hidelinks]{hyperref}

\hypersetup{
  pdftitle={EmbodiedMonitor: Can Embodied Agents Reliably Assess Their Own Performance?},
  pdfauthor={Xueyuan Yang, Yuxi Wang, Ruizhe Chen, Zhenliang Zhang, Yujia Peng}
}
\title{\LARGE\bf EmbodiedMonitor: Can Embodied Agents Reliably Assess Their Own Performance?}
\author{Xueyuan Yang$^{1,\dagger}$, Yuxi Wang$^{1,2,\dagger,*}$, Ruizhe Chen$^{3}$, Zhenliang Zhang$^{1,*}$, and Yujia Peng$^{1,2,4,*}$%
\thanks{$^{\dagger}$Equal contribution.}%
\thanks{$^{*}$Corresponding authors: Yuxi Wang ({\tt\small wangyuxi@bigai.ai}), Zhenliang Zhang ({\tt\small zzlyw10@gmail.com}), and Yujia Peng ({\tt\small yujia\_peng@pku.edu.cn}).}%
\thanks{$^{1}$State Key Laboratory of General Artificial Intelligence, Beijing Institute for General Artificial Intelligence (BIGAI), Beijing 100080, China.}%
\thanks{$^{2}$School of Psychological and Cognitive Sciences and Beijing Key Laboratory of Behavior and Mental Health, Key Laboratory of Machine Perception (Ministry of Education), Peking University, Beijing 100871, China.}%
\thanks{$^{3}$School of Electronics Engineering and Computer Science, Peking University, Beijing 100871, China.}%
\thanks{$^{4}$Institute for Artificial Intelligence, Peking University, Beijing 100871, China.}%
}
\begin{document}
\maketitle
\thispagestyle{plain}
\begin{abstract}
Appropriate trust in human--robot collaboration depends on embodied agents' ability to assess their own performance accurately and communicate uncertainty reliably. The dynamic nature of embodied tasks and the uncertainty of execution outcomes require self-assessment throughout the entire execution process: agents must continually judge current progress, recognize the effects of individual actions, and update their expectations of the final outcome accordingly. We introduce EmbodiedMonitor, a benchmark for self-monitoring before, during, and after task execution. The benchmark spans 39 high-fidelity indoor scenes, 152 object categories, and 41,321 task instances across four levels, incorporating controlled perturbations such as action failures and state disruptions. Along the same execution trajectory, the benchmark precisely aligns four core types of self-assessment across the pre-execution, online, and post-execution stages---current-progress judgments, process-change judgments, final-outcome forecasts, and post-execution retrospective assessments---with the underlying environmental ground truth, jointly evaluating judgment accuracy and confidence calibration. Closed-loop evaluation of eight widely used multimodal models reveals that, while current frontier models exhibit some self-assessment capabilities, progress overestimation, overconfidence, false progress reports, and missed regressions remain widespread. These findings indicate that current models cannot yet support reliable trust calibration in human--robot collaboration. EmbodiedMonitor provides a standardized benchmark and empirical evidence for systematically diagnosing the reliability limits of embodied self-assessment and advancing reliable self-monitoring methods.
\end{abstract}
\section{Introduction}
\label{sec:introduction}

A robot's self-assessment reports on its capability limits and task performance provide a key basis for calibrating human--robot trust and enabling safe collaboration~\cite{norton2022}. Behavioral experiments show that capability self-assessments provided before execution can effectively help operators weigh autonomous execution against human takeover~\cite{conlon2022}. Accurate self-assessment and reliable confidence reporting are therefore fundamental to trustworthy human--robot collaboration.

\begin{figure}[!t]
  \centering
  \includegraphics[width=\columnwidth]{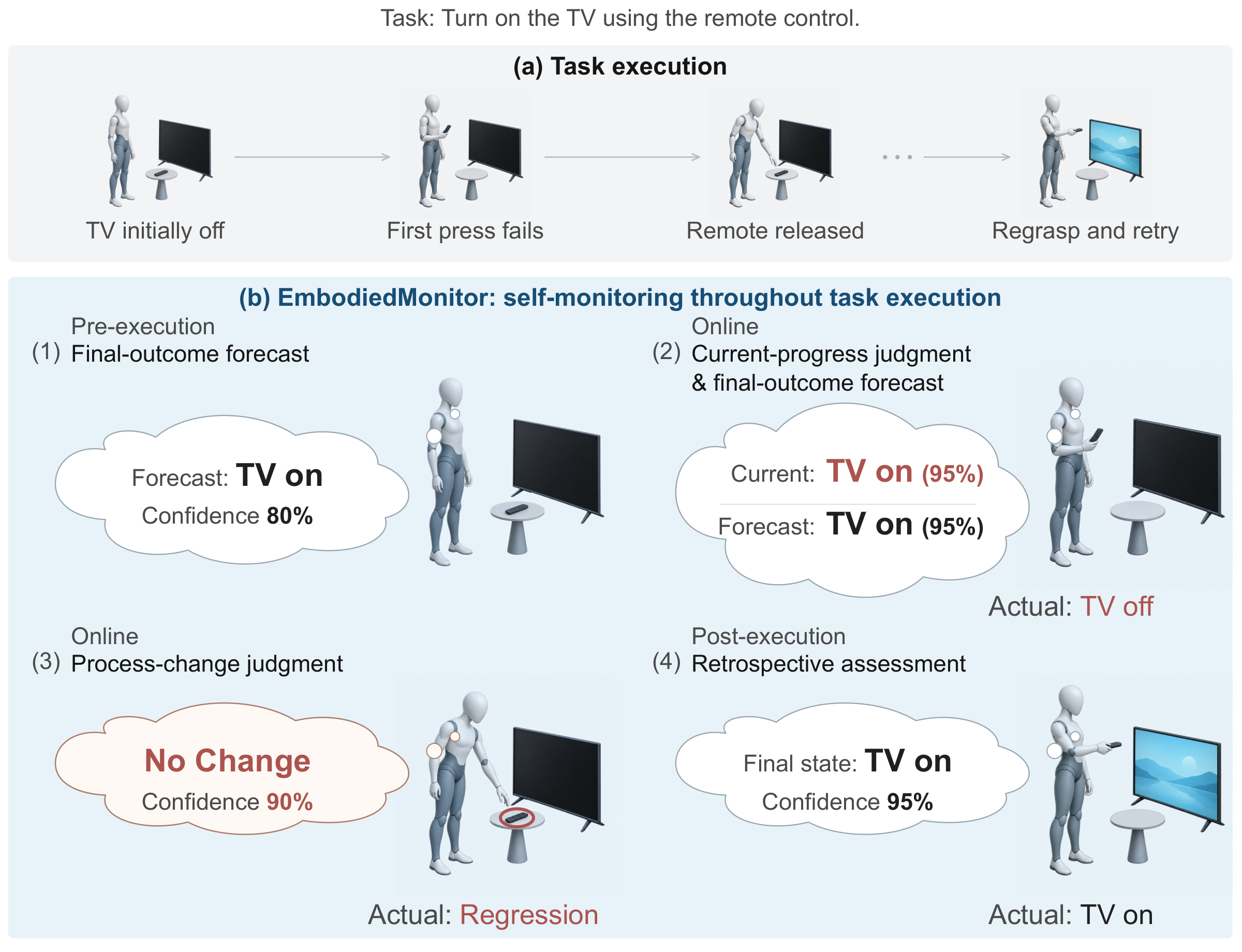}
  \caption{Self-monitoring throughout task execution. An agent may ultimately complete a task while missing a regression in task conditions with high confidence during execution.}
  \label{fig:overview}
\end{figure}

However, the dynamic nature of physical environments and the uncertainty of action outcomes mean that embodied self-monitoring cannot stop at a static judgment at a single point in time. Initial expectations formed before execution must be continually tested and revised as the environment changes and action feedback becomes available~\cite{conlon2024}: during execution, agents must assess task progress, use observations and feedback to judge whether actions have produced their intended effects, and update their final-outcome forecasts accordingly; after execution, self-assessment shifts to reviewing what was actually accomplished~\cite{frasca2020}. Self-monitoring thus spans the complete closed loop before, during, and after execution, characterizing agents' evolving understanding of task states and expected outcomes across execution stages (Fig.~\ref{fig:overview}).

Existing embodied AI benchmarks primarily evaluate agent performance through objective execution outcomes, such as task success rates and subgoal completion~\cite{embodiedbench2025,eai2024}, while overlooking whether agents' internal self-assessments align with objective facts. Recent studies have begun to investigate self-assessment and uncertainty in embodied settings: \emph{Uncertainty in Action} evaluates confidence calibration and failure prediction against perception correctness and final task success or failure~\cite{yu2025}; \emph{SCOPER} examines judgments of object states, action effects, and task progress through execution-linked question answering~\cite{wei2026scoper}. However, these approaches either focus on perception correctness and final success or failure without tracking changes in task progress, or assess the accuracy of semantic judgments without incorporating confidence calibration. A systematic joint evaluation that connects pre-execution forecasts, online monitoring, and post-execution retrospective assessments along the same trajectory remains lacking.

To address this gap, we develop EmbodiedMonitor on TongSIM~\cite{tongsim2025} as a benchmark for self-monitoring throughout task execution. Through unified observation and action interfaces, EmbodiedMonitor continuously collects four core types of self-assessment at task and action granularities along the same trajectory: current-progress judgments, final-outcome forecasts before and during execution, post-execution retrospective assessments, and process-change judgments. These reports are precisely aligned with the underlying environmental ground truth to jointly evaluate judgment accuracy and confidence calibration within a unified framework. To cover diverse observation conditions and goal dependencies, the benchmark spans 39 high-fidelity indoor scenes, 152 object categories, and 5,577 object instances, from which we construct a library of 41,321 task instances organized into four levels: bodily actions, navigation and localization, single-goal interactions, and composite tasks. We further introduce controlled perturbations---including action failures, incomplete execution, object displacement, and disruption of established conditions---that cause actual state transitions to deviate explicitly from intended action effects. These perturbations test whether agents can promptly update their self-assessments using environmental observations and feedback, rather than inferring progress solely from historical action sequences.

Using EmbodiedMonitor, we conduct closed-loop evaluations of eight widely used multimodal models, revealing a divergence across core embodied self-monitoring capabilities. As execution proceeds, final-outcome forecasts progressively approach actual outcomes, whereas overestimation of current progress grows substantially in later stages, and overconfidence remains prevalent across judgment types. Models are also highly prone to reporting unchanged states as progress and missing regressions of established conditions; these errors further skew final-outcome expectations and induce unwarranted increases in confidence. Nevertheless, reported confidence retains some ability to distinguish correct from incorrect judgments. These findings indicate that current models remain insufficient to support reliable trust calibration in human--robot collaboration, while providing empirical evidence for selective verification and human intervention based on self-assessment signals.

Our main contributions are as follows:
\begin{enumerate}
  \item \textbf{A protocol for evaluating self-monitoring along a single execution trajectory.} We use simulator states and action logs to score four types of self-assessment across execution stages and assess the calibration of their reported confidence.
  \item \textbf{A large-scale, multilevel benchmark and a closed-loop execution baseline.} Built on TongSIM, EmbodiedMonitor spans 39 high-fidelity indoor scenes, 152 object categories, 5,577 object instances, and 41,321 task instances across four task levels. Controlled action failures and state disruptions challenge the assumption of ideal action execution, while a closed-loop execution baseline provides unified observation and action interfaces.
  \item \textbf{Capability limits and typical failure modes of embodied self-monitoring.} Closed-loop evaluation of eight widely used multimodal models reveals that, while current frontier models exhibit some self-assessment capabilities, progress overestimation, overconfidence, false progress reports, and missed regressions remain widespread. These findings indicate that current models cannot yet support reliable trust calibration in human--robot collaboration.
\end{enumerate}
\section{Related Work}
\label{sec:related-work}

\subsection{Performance Evaluation in Embodied Benchmarks}

Existing embodied AI benchmarks primarily evaluate agent capabilities through objective external execution performance. Their core metrics include overall task success rates~\cite{shridhar2020,shridhar2021,padmakumar2022}, goal-condition satisfaction, and subtask completion~\cite{srivastava2022,li2023,szot2021}. To further identify the sources of failure, recent process-diagnostic benchmarks extend evaluation to objective error attribution in intermediate stages such as planning, navigation, and manipulation~\cite{embodiedbench2025,eai2024}. Although these benchmarks provide objective references for task outcomes and execution processes, they remain limited to assessing how well an agent actually performs, without examining whether its judgments of progress, action effects, and final outcomes align with objective facts.

\subsection{Agent Self-Monitoring and Confidence}

Agent self-monitoring concerns internal assessments of capability limits and task performance, while confidence quantifies uncertainty in these judgments~\cite{norton2022,steyvers2026}. Systematic evaluation of self-assessment involves two core dimensions: whether the content of a judgment aligns with objective facts (judgment accuracy), and whether reported confidence matches empirical correctness rates (confidence calibration)~\cite{guo2017}.

Model self-assessment and uncertainty estimation have been widely studied in static, single-turn tasks such as natural language and visual question answering. Work on large language models examines self-assessment mechanisms for predicting whether a question can be answered correctly beforehand and evaluating answer correctness afterward~\cite{kadavath2022}, and improves confidence calibration through prompting, fine-tuning, or repeated sampling and aggregation~\cite{lin2022,tian2023,xiong2024}. Research on visual question answering extends self-assessment to multimodal inputs, comparing confidence signals and calibration methods against answer correctness~\cite{ding2025}. However, these studies are limited to single-turn decision settings with static environmental states and do not account for continuous state transitions or multistep observation feedback.

In contrast, long-horizon interactive tasks typically involve multistep dependencies and environmental nonstationarity. Initial expectations formed before execution can rapidly become invalid as states evolve and perception and execution errors accumulate, requiring agents to dynamically recalibrate their beliefs about task progress and final outcomes based on a continuous stream of observational evidence~\cite{israelsen2025famsec,yu2025}. In human--robot collaboration, communicating such self-assessments to human operators provides a key basis for weighing autonomous execution against human intervention and establishing appropriate trust~\cite{norton2022,warmsley2025}; behavioral experiments also show that robot capability self-assessments can effectively improve operators' reliance decisions and team performance~\cite{conlon2022}. At the conceptual level, Frasca et al.\ propose a three-stage framework encompassing pre-execution capability prediction, online progress assessment, and post-execution retrospective assessment, initially validated through task dialogues~\cite{frasca2020}.

Recent studies have begun to quantify self-assessment and uncertainty in long-horizon interactive tasks; their evaluation coverage and input modalities are compared in Table~\ref{tab:related-self-assessment}. However, three key aspects remain underexplored. First, simplified perceptual modalities: \emph{Agent-BRACE}~\cite{singh2026brace} and \emph{Outcome-Grounded}~\cite{hathidara2026} rely on text-based observations (e.g., in TextWorld~\cite{cote2018textworld}, ALFWorld~\cite{shridhar2021}, and WebShop~\cite{yao2022webshop}) and do not directly examine agents' ability to verify action effects and state changes from first-person images. Second, narrow self-assessment targets: \emph{Uncertainty in Action}~\cite{yu2025} focuses on calibration with respect to perception correctness and final task success or failure; \emph{Outcome-Grounded}~\cite{hathidara2026} examines final-success probabilities before and during execution; and \emph{Agent-BRACE}~\cite{singh2026brace} emphasizes calibration of subgoal-state beliefs. These studies do not jointly examine the evolution of multiple self-assessment types or their relationships across granularities. Third, separate treatment of judgments and calibration: although \emph{SCOPER}~\cite{wei2026scoper} examines current progress, action effects, and retrospective judgments in visual environments, it omits confidence calibration. Jointly evaluating the accuracy and confidence calibration of multiple self-assessment types along the same trajectory---spanning the pre-execution, online, and post-execution stages in a closed loop grounded in first-person visual observations and structured state feedback---therefore remains an important research gap.

\begin{table*}[!t]
  \caption{Environments, input modalities, and evaluation coverage of embodied self-assessment studies. Acc.: judgment accuracy; Cal.: confidence calibration; $\checkmark$: covered; ---: not covered.}
  \label{tab:related-self-assessment}
  \centering
  \footnotesize
  \setlength{\tabcolsep}{1.5pt}
\renewcommand{\arraystretch}{1.12}
\renewcommand{\tabularxcolumn}[1]{m{#1}}
  \begin{tabularx}{\textwidth}{@{}>{\raggedright\arraybackslash}m{0.16\textwidth}>{\raggedright\arraybackslash}m{0.16\textwidth}>{\centering\arraybackslash}m{0.10\textwidth}*{10}{>{\centering\arraybackslash}X}@{}}
    \toprule
    & & & \multicolumn{2}{c}{Pre-execution} & \multicolumn{6}{c}{Online} & \multicolumn{2}{c}{Post-execution} \\
    \cmidrule(lr){4-5}\cmidrule(lr){6-11}\cmidrule(lr){12-13}
    & & & \multicolumn{2}{c}{\shortstack{Final-outcome\\forecast}} & \multicolumn{2}{c}{\shortstack{Current-progress\\judgment}} & \multicolumn{2}{c}{\shortstack{Process-change\\judgment}} & \multicolumn{2}{c}{\shortstack{Final-outcome\\forecast}} & \multicolumn{2}{c}{\shortstack{Retrospective\\assessment}} \\
    \cmidrule(lr){4-5}\cmidrule(lr){6-7}\cmidrule(lr){8-9}\cmidrule(lr){10-11}\cmidrule(lr){12-13}
    \textbf{Work} & \textbf{Environment} & \textbf{Input} & Acc. & Cal. & Acc. & Cal. & Acc. & Cal. & Acc. & Cal. & Acc. & Cal. \\
    \midrule
    \shortstack[l]{Uncertainty\\in Action} & Minecraft~\cite{johnson2016malmo} & \mbox{Image + text} & --- & --- & --- & --- & --- & --- & $\checkmark$ & $\checkmark$ & --- & --- \\
\addlinespace[3pt]
    SCOPER & \shortstack[l]{BEHAVIOR~\cite{srivastava2022}\\ProcTHOR~\cite{deitke2022procthor}} & \mbox{Image + text} & --- & --- & $\checkmark$ & --- & $\checkmark$ & --- & --- & --- & $\checkmark$ & --- \\
\addlinespace[3pt]
    Agent-BRACE & \shortstack[l]{TextWorld~\cite{cote2018textworld}\\ALFWorld~\cite{shridhar2021}} & Text & --- & --- & $\checkmark$ & $\checkmark$ & --- & --- & --- & --- & --- & --- \\
\addlinespace[3pt]
    Outcome-Grounded & \shortstack[l]{ALFWorld~\cite{shridhar2021}\\TextWorld~\cite{cote2018textworld}\\WebShop~\cite{yao2022webshop}} & Text & $\checkmark$ & $\checkmark$ & --- & --- & --- & --- & $\checkmark$ & $\checkmark$ & --- & --- \\
\addlinespace[3pt]
\midrule
    \textbf{EmbodiedMonitor}\newline\textbf{(ours)} & \textbf{TongSIM}~\cite{tongsim2025} & \mbox{Image + text} & $\checkmark$ & $\checkmark$ & $\checkmark$ & $\checkmark$ & $\checkmark$ & $\checkmark$ & $\checkmark$ & $\checkmark$ & $\checkmark$ & $\checkmark$ \\
    
\bottomrule
  \end{tabularx}
\end{table*}

To address this gap, EmbodiedMonitor leverages verifiable environmental ground truth within a closed loop combining first-person visual observations with structured egocentric and visible-object state records to jointly evaluate judgment accuracy and confidence calibration across four core types of self-assessment along the same trajectory (Table~\ref{tab:related-self-assessment}). In addition, controlled perturbations such as action failures and state disruptions cause actual states to deviate explicitly from intended action effects, rigorously testing whether agents can dynamically update their self-assessments based on actual observations and feedback. EmbodiedMonitor thereby establishes a systematic evaluation paradigm spanning the entire execution loop, providing an empirical foundation for diagnosing reliability bottlenecks in embodied self-assessment and calibrating human--robot trust.
\section{Benchmark and Evaluation Protocol}
\label{sec:benchmark}

\subsection{Task Design and Condition Representation}
\label{sec:task-design}

\begin{figure*}[!t]
    \centering
    \includegraphics[width=\textwidth,height=0.55\textheight,keepaspectratio]{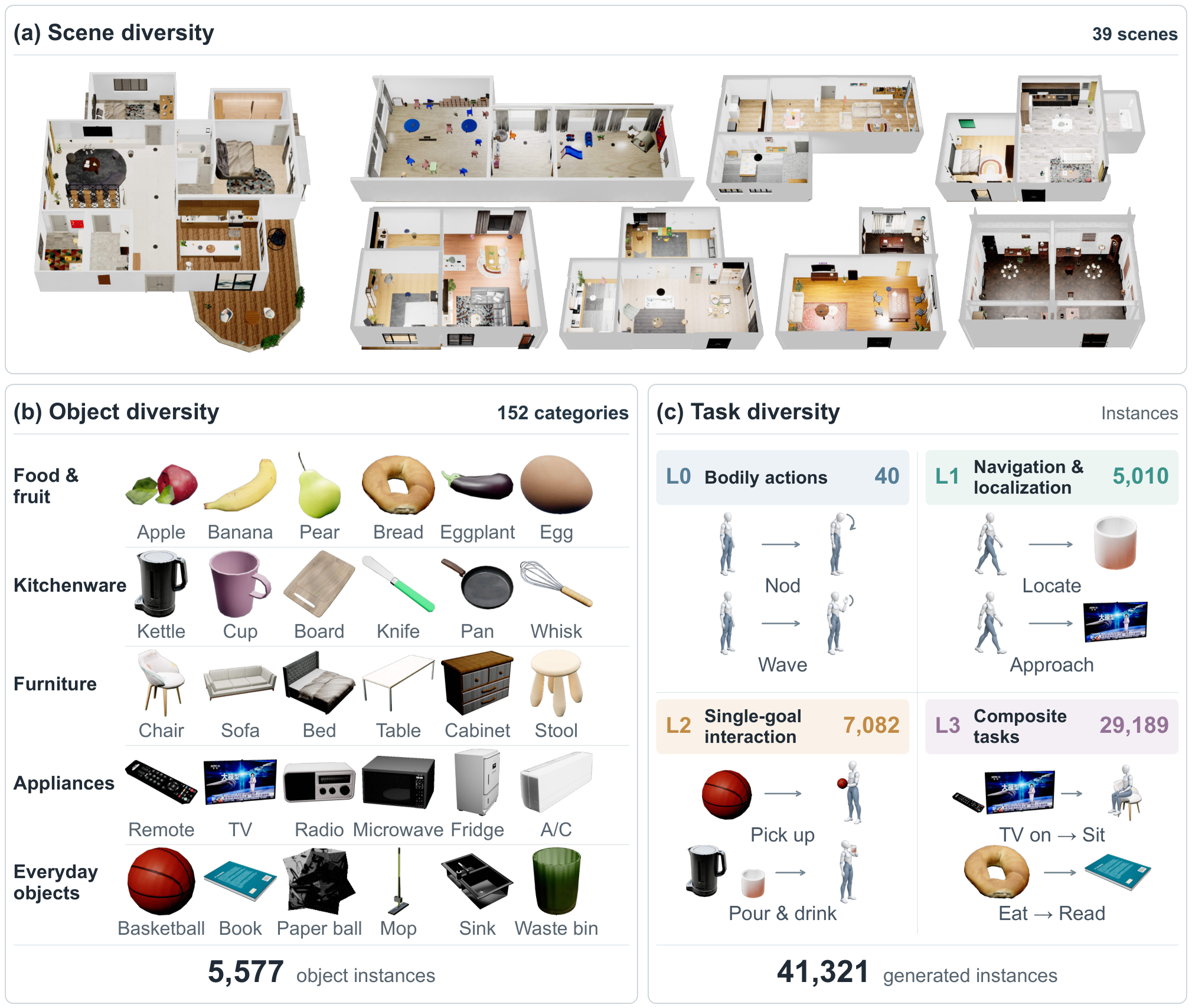}
    \caption{Benchmark composition and diversity across scenes, objects, and tasks: (a)~representative layouts from 39 indoor environments, (b)~5,577 object instances spanning 152 categories across five functional groups, and (c)~41,321 task instances stratified into four complexity levels (L0--L3).}
    \label{fig:diversity}
\end{figure*}

Built on TongSIM~\cite{tongsim2025}, EmbodiedMonitor provides a large-scale library of embodied tasks spanning diverse interaction scenarios. It contains 164 representative task designs across 86 task categories, yielding 41,321 instances in 39 indoor scenes, with 5,577 object instances from 152 categories (Fig.~\ref{fig:diversity}). To encompass interactions ranging from basic actions to multi-goal coordination, tasks are organized by complexity into four levels: bodily actions (L0), navigation and localization (L1), single-goal interactions (L2), and composite tasks (L3). These levels respectively involve executing specified bodily actions, searching for and approaching targets, manipulating a single object, and fulfilling ordering and joint-satisfaction constraints across multiple goals. The environments include kitchens, dining rooms, living rooms, bedrooms, bathrooms, studies, office areas, and children's activity spaces. Objects include food and kitchenware (e.g., bread, cups, and kettles, supporting eating, drinking, and pouring), furniture (e.g., sofas and chairs, supporting sitting and lying down), appliances (e.g., televisions and air conditioners, supporting device-state control), and cleaning equipment (e.g., mops and trash bins, supporting stain removal and waste disposal). Combinations of target objects, initial agent poses, environmental lighting, and initial device states further generate diverse task instances with varying observation conditions and interaction starting points.

To characterize execution progress consistently across tasks, we formalize each task as a set of verifiable conditions covering object and agent states, spatial relations, possession relations, and prescribed behaviors, with explicit dependencies and ordering constraints. This set is divided into \emph{task requirements}, which constitute the goals used to evaluate task completion, and \emph{execution prerequisites}, which are the preparatory states needed to satisfy those requirements. For state conditions (e.g., keeping a television on), the evaluator continuously monitors their current truth values; previously satisfied conditions can become invalid after subsequent state changes. For prescribed behaviors (e.g., waving), it retains a satisfaction record once the behavior has been verified as completed. By comparing satisfaction states in the current monitoring set before and after an action, the evaluator defines three categories of single-step process change: \emph{Progress} when conditions become newly satisfied, \emph{No Change} when no satisfaction states change, and \emph{Regression} when previously satisfied conditions are lost. These categories assess models' ability to recognize single-step state transitions. For example, in the task ``put the key on the table,'' holding the key is a prerequisite for placement: picking it up constitutes progress, and dropping it before placement constitutes regression. Once placement is complete, the possession prerequisite is removed from the current monitoring set because it is no longer needed.

To simulate execution uncertainty in physical interaction and remove the assumption of deterministic state transitions in idealized simulation, the benchmark introduces 17 controlled perturbations across six categories (Table~\ref{tab:perturbations}). During execution, perturbations are triggered only at eligible steps where the current action and task state satisfy specific preconditions; at such moments, one applicable perturbation is selected with $80\%$ probability (at most one per step). Within a single episode, L0--L1 and L2--L3 tasks receive at most two and three perturbations, respectively, and each condition is perturbed at most once per episode, keeping perturbations sparse in practice while preserving task recoverability. Rather than returning explicit interface error messages, perturbation effects manifest directly in the subsequent environmental state and observations. In most cases, these deviations are immediately observable from the next first-person image and structured state records; in a few visually subtle or partially occluded cases (e.g., unresponsive door opening or closing when only part of the door lies within the current field of view), they reflect realistic perceptual ambiguity. Together, these perturbations allow us to assess whether agents verify action effects from actual observations, detect the loss of previously satisfied conditions, and update their progress judgments and confidence accordingly.

\begin{table*}[!t]
    \caption{Controlled perturbation settings. Six categories encompass 17 controlled perturbations.}
    \label{tab:perturbations}
    \centering
    \small
    \begin{tabularx}{\textwidth}{@{}>{\raggedright\arraybackslash}p{0.21\textwidth}>{\raggedright\arraybackslash}X>{\raggedright\arraybackslash}p{0.24\textwidth}@{}}
        \toprule
        Perturbation category & Setting & Evaluation focus \\
        \midrule
        Unresponsive actions~(7) & Suppress the intended effects of device interaction, door opening or closing, pointing, mopping, sitting down, standing up, and climbing. & Verify action effects from actual outcomes. \\
        Incomplete execution~(3) & Stop navigation early; execute only half of a requested fixed-distance movement or fixed-angle rotation. & Distinguish partial execution from meeting the task criterion. \\
        Object displacement~(2) & Drop a held target object; offset placement so that an object lands outside a trash bin. & Assess possession and placement outcomes accurately. \\
        Loss of interaction readiness~(3) & Release a tool, food item, or remote control before use or interaction, or turn away to invalidate effective reach. & Detect the loss of established manipulation conditions. \\
        Loss of proximity~(1) & Move the agent outside the distance range specified by the task before an operation. & Continuously verify spatial conditions. \\
        Lighting-state reversal~(1) & After the lights have been turned off, intercept a request to sit or lie down and turn the lights back on. & Detect the invalidation of a previously satisfied state. \\
        \bottomrule
    \end{tabularx}
\end{table*}

\subsection{Agent Interaction and Full-Cycle Self-Assessment}

\begin{figure*}[!t]
    \centering
    \includegraphics[width=0.92\textwidth]{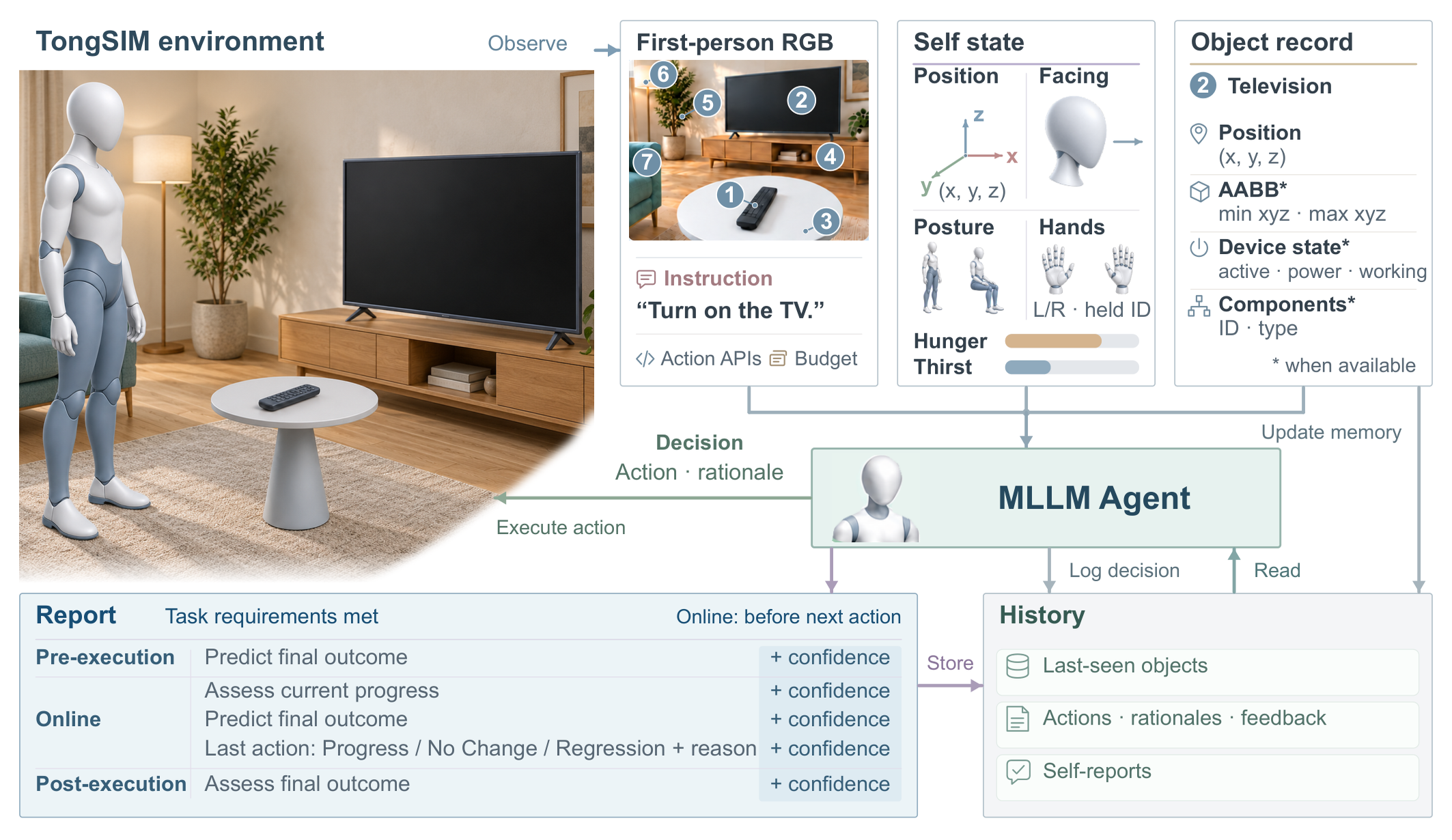}
    \caption{Closed-loop interaction and full-cycle self-assessment pipeline. At each round, the MLLM agent combines an ID-annotated first-person RGB image, structured self states, object records, and history to emit stage-specific self-assessment reports with confidence scores prior to action selection.}
    \label{fig:agent}
\end{figure*}

Agents execute tasks through unified observation and action interfaces and generate self-assessment reports throughout execution (Fig.~\ref{fig:agent}). In each decision round, the agent receives the task instruction, descriptions of available actions, the remaining decision budget, current observations, and historical information. The current observation comprises a single first-person image alongside structured text specifying the agent's own state (position, orientation, posture, possession, hunger, and thirst) and the identifiers, spatial locations, and available states of visible objects. To align visual entities with these structured object records, visible objects in the first-person image are overlaid with numeric identifiers matching their entries in the observation text (Fig.~\ref{fig:agent}). Historical information comprises execution records and object memory. The former records all executed actions in the episode, decision rationales, pre-action agent states, and past self-assessment reports (including current-progress judgments, process-change judgments, final-outcome forecasts, and their associated confidence scores). The latter retains the most recent observations of objects that are currently out of view.

In each decision round, the agent first assesses the task state and generates self-assessment reports from the current observation and historical information. It then selects at most one action from a space of 42 interfaces (Table~\ref{tab:actions}) or voluntarily terminates the task, along with a decision rationale. After an action is executed, the environment returns a new observation and execution feedback. The new observation updates object memory, while the action, feedback, and self-assessment results are added to the execution history. Together, these serve as inputs to the next round for assessing the single-step action effect, updating task progress, and continuing decision-making. This closed loop repeats until the agent voluntarily terminates or exhausts its decision budget.

\begin{table*}[!t]
    \caption{Agent action space. The 42 interfaces are grouped by function, with representative calls listed.}
    \label{tab:actions}
    \centering
    \small
    \begin{tabularx}{\textwidth}{@{}>{\raggedright\arraybackslash}p{0.17\textwidth}>{\raggedright\arraybackslash}X>{\raggedright\arraybackslash}p{0.34\textwidth}@{}}
        \toprule
        Action category & Supported operations & Representative interfaces \\
        \midrule
        Navigation and localization & Move to a target object, component, or specified location. & \texttt{MoveToObject}, \texttt{MoveToLocation} \\
        Orientation and gaze & Face a target, turn in place, and adjust gaze. & \texttt{TurnToObject}, \texttt{TurnInDegree}, \texttt{LookAtLocation} \\
        Hand and object manipulation & Reach out, retract the hand, and pick up or place objects. & \texttt{HandReach}, \texttt{TakeObject}, \texttt{PutDownToLocation} \\
        Door and device control & Open or close doors, operate devices, and toggle environmental lighting. & \texttt{OpenDoor}, \texttt{Interact}, \texttt{TurnEnvironmentLightsOff} \\
        Eating and pouring & Eat or drink a held item, and pour water into a cup. & \texttt{EatOrDrink}, \texttt{PourWater} \\
        Cleaning and washing & Mop floors, wash the face or hands, and wash a held object. & \texttt{MopFloor}, \texttt{WashFace}, \texttt{WashObjectInHand} \\
        Posture and climbing & Sit down, lie down, stand up, climb, and climb down. & \texttt{SitDownToObject}, \texttt{SleepDown}, \texttt{ClimbObject} \\
        Gestures and action performance & Point, wave, raise a hand, nod, shake the head, and perform facial expressions. & \texttt{PointAtObject}, \texttt{WaveHand}, \texttt{PlayAnimation} \\
        Reading and waiting & Read and wait for a specified duration. & \texttt{ReadBook}, \texttt{Wait} \\
        \bottomrule
    \end{tabularx}
\end{table*}

Agents generate four core types of self-assessment at different stages throughout execution, reporting confidence $q$ for each. Before execution, they forecast which task requirements will ultimately be satisfied within the decision budget. During execution, they jointly produce current-progress judgments, process-change judgments, and online final-outcome forecasts, respectively reporting the currently satisfied requirements, the task-relevant changes from the latest step and their supporting evidence, and revised completion expectations. After execution, the same agent uses the final observation and complete execution history to retrospectively assess which requirements were actually satisfied. We collectively refer to pre-execution and online final-outcome forecasts, together with post-execution retrospective assessments, as \emph{final-outcome judgments}, capturing the evolution of the agent's understanding of the final result. For current-progress and all final-outcome judgments, $q$ denotes the subjective probability that the reported set exactly matches the actually satisfied requirements, with no omissions or additions. For process-change judgments, $q$ denotes the subjective probability that the reported change category is correct.

\subsection{Evaluation Protocol}

To automate objective verification of self-assessment throughout execution, the evaluator directly reads the underlying simulator states and action logs to generate ground-truth task progress at each time point and single-step process-change labels. This ground truth is stored separately on the evaluation side to isolate it from the agent. Online current-progress and process-change judgments are verified against the actual state at the corresponding time point and the single-step state change, respectively. Pre-execution and online final-outcome forecasts, together with post-execution retrospective assessments, are all compared against actual task completion at termination. An individual report is considered correct if its reported set matches the ground truth exactly or its change category agrees with the ground truth, as applicable. Alongside judgment accuracy, we use expected calibration error (ECE) to measure the alignment between reported confidence $q$ and empirical accuracy~\cite{guo2017}. Confidence is divided into 10 equal-width bins, and ECE is defined as

\[
\mathrm{ECE}=\sum_{b=1}^{10}p_b\left|\mathrm{acc}(b)-\mathrm{conf}(b)\right|,
\]

where $p_b$ is the fraction of the total report weight assigned to bin $b$, and $\mathrm{acc}(b)$ and $\mathrm{conf}(b)$ are the weighted accuracy and weighted mean confidence of reports in that bin, respectively. For each model and judgment type, each task's weight is divided equally among its valid online reports, with equal weight given to all contributing tasks. For stage-wise ECE, this weighting is applied within each stage.
\section{Experiments}
\label{sec:experiments}

\begin{figure*}[!t]
  \centering
  \includegraphics[width=0.90\textwidth]{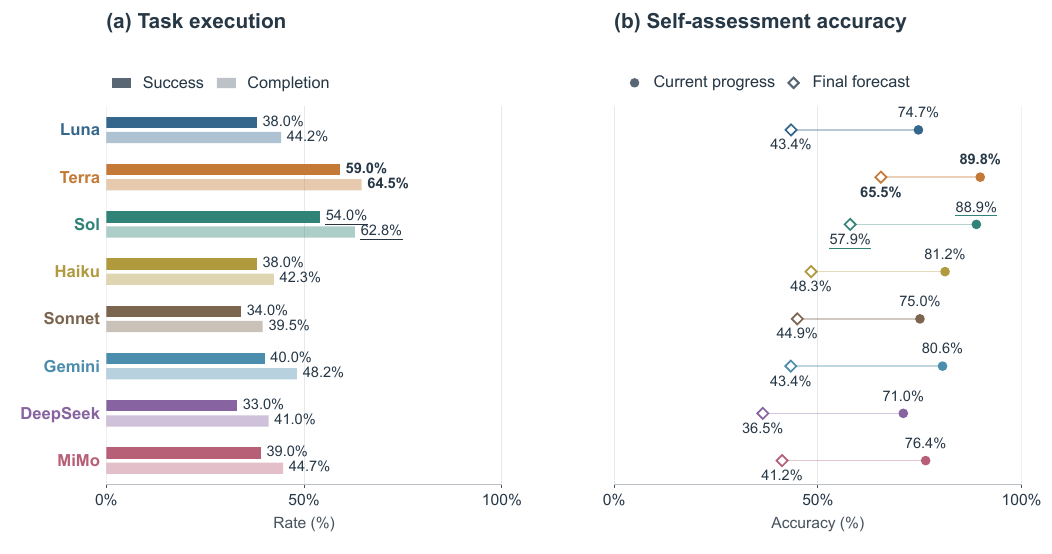}
  \caption{Task execution and self-assessment performance. (a) Task success rate and mean completion. (b) Overall accuracy of current-progress judgments and online final-outcome forecasts. All values are percentages. Bold and underlined values denote the best and second-best results for each metric, respectively.}
  \label{fig:execution}
\end{figure*}

\begin{figure*}[!t]
  \centering
  \includegraphics[width=\textwidth,height=0.74\textheight,keepaspectratio]{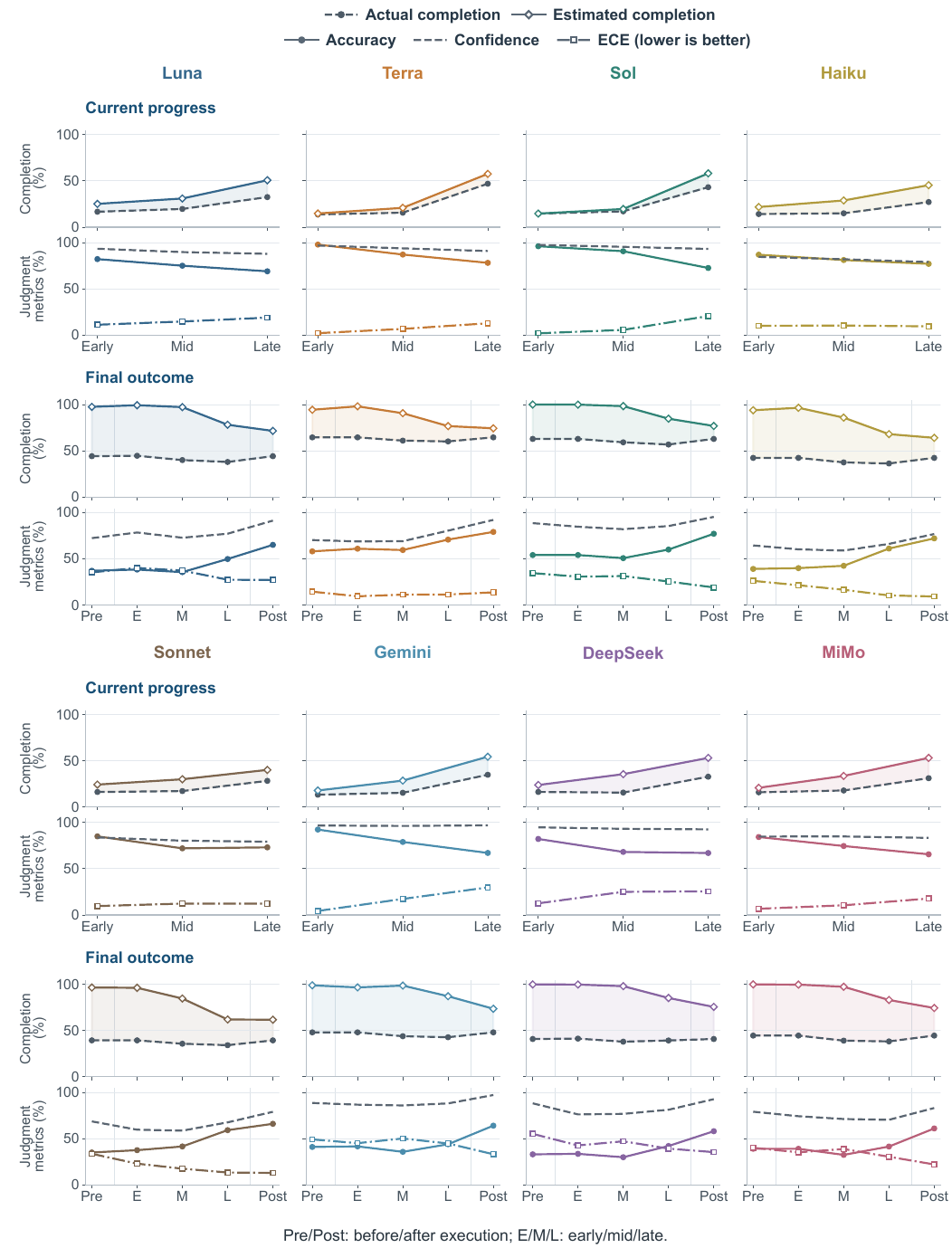}
  \caption{Self-assessment across execution stages. The upper and lower groups show current-progress and final-outcome judgments, respectively. In each group, the upper row shows actual and estimated completion, with shading indicating overestimation, and the lower row shows judgment accuracy, mean confidence, and ECE. Pre and Post denote the pre-execution and post-execution stages; online reports are divided approximately into thirds by order, corresponding to early (E), middle (M), and late (L) stages. Some short trajectories cannot populate all three stages, so sample sizes differ across stages; actual task completion is averaged over the corresponding trajectory subset at each stage. All metrics are percentages.}
  \label{fig:monitoring}
\end{figure*}

\subsection{Experimental Setup}

To cover different levels of task complexity, we sample 100 tasks from the benchmark pool using level-specific quotas. The shared test set spans 31 scenes and includes 15 bodily-action tasks (L0), 25 navigation and localization tasks (L1), 25 single-goal interaction tasks (L2), and 35 composite tasks (L3). Controlled perturbations are enabled according to the protocol in Section~\ref{sec:task-design}. We evaluate eight widely used multimodal models: gpt-6-luna (Luna), gpt-6.1-terra (Terra), gpt-6.1-sol (Sol), claude-haiku-5-5 (Haiku), claude-sonnet-5-5 (Sonnet), gemini-3.8-flash (Gemini), deepseek-v4.1-flash-1017 (DeepSeek), and mimo-v2.6-pro (MiMo). Across the 800 task executions, a mean of 0.924 perturbations were applied per execution, and $49.1\%$ of executions had no perturbation. All models use the same agent framework and observation--action interfaces. The decision budget is 2--3 rounds for L0 tasks and 8, 10, and 15 rounds for L1, L2, and L3 tasks, respectively. Reasoning effort is set to \texttt{high}, with a maximum output of 16,384 tokens per call. Given the substantial computational cost of high-effort reasoning across eight models in multiround visual closed-loop interaction, this stratified test set spans all task levels within a practical evaluation budget while still yielding thousands of step- and stage-level verification samples. The remaining benchmark instances serve as a contamination-resistant reserve set and a training data pool.

Our experiments strictly follow the interaction and self-assessment protocol in Section~\ref{sec:benchmark}, collecting the four core types of self-assessment and their confidence scores before, during, and after execution along the same trajectory, with automated verification against the underlying simulator ground truth. Evaluation covers execution performance and self-monitoring. Objective execution performance is measured by task success rate (the proportion of tasks in which all requirements are satisfied) and mean completion (the average proportion of satisfied requirements per task). Self-monitoring is assessed jointly through judgment accuracy and confidence calibration.

\begin{figure*}[!t]
  \centering
  \includegraphics[width=\textwidth,height=0.59\textheight,keepaspectratio]{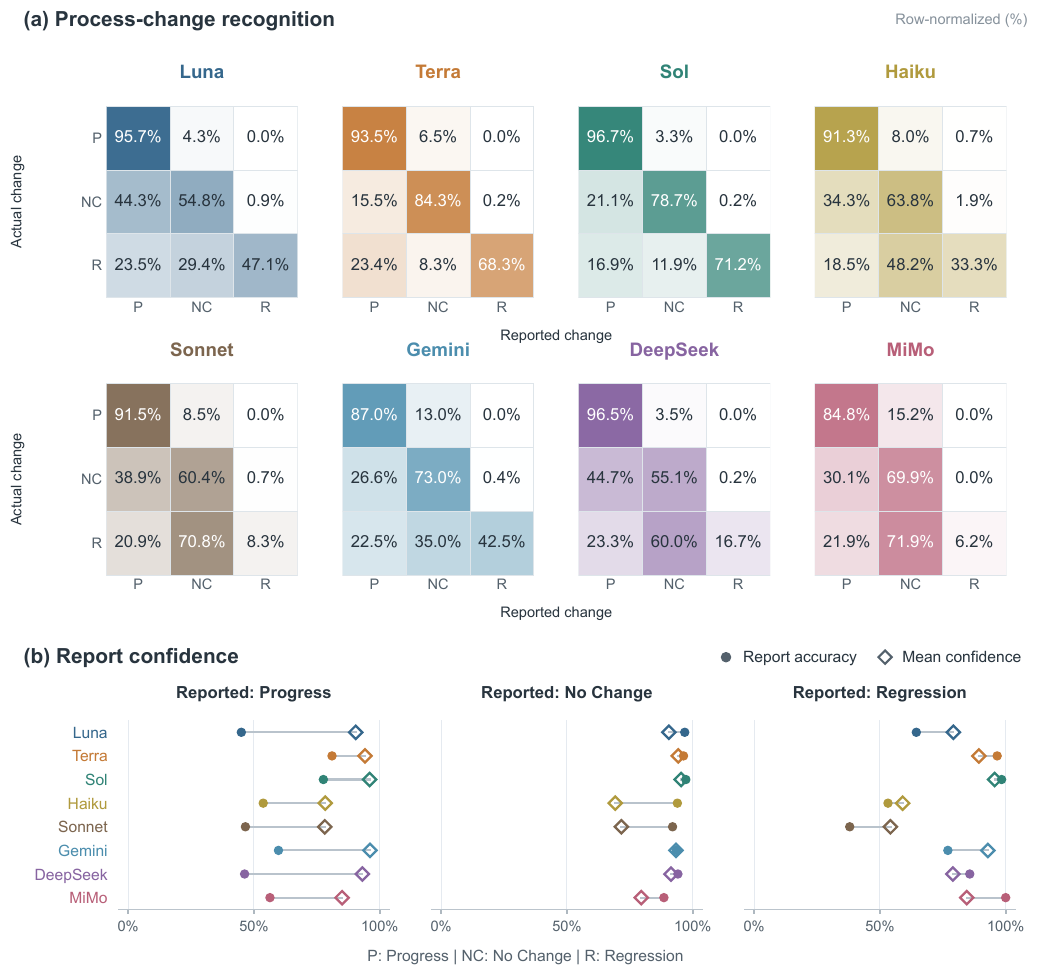}
  \caption{Change recognition and report confidence. (a) Row-normalized confusion matrices of process-change judgments. (b) Accuracy (filled circles) and mean confidence (open diamonds) by reported category.}
  \label{fig:process}
\end{figure*}

\begin{figure}[!t]
  \centering
  \includegraphics[width=\columnwidth]{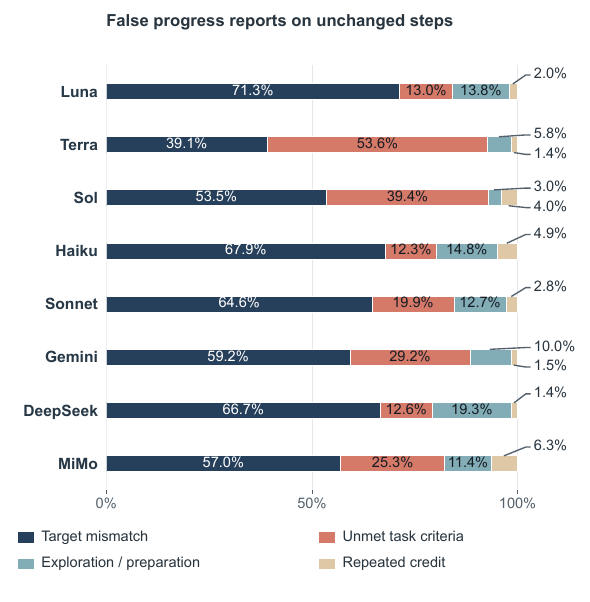}
  \caption{Error-attribution breakdown of no-change steps misreported as progress across the eight evaluated models, grouped by four failure categories.}
  \label{fig:errors}
\end{figure}

\begin{figure*}[!t]
  \centering
  \includegraphics[width=\textwidth]{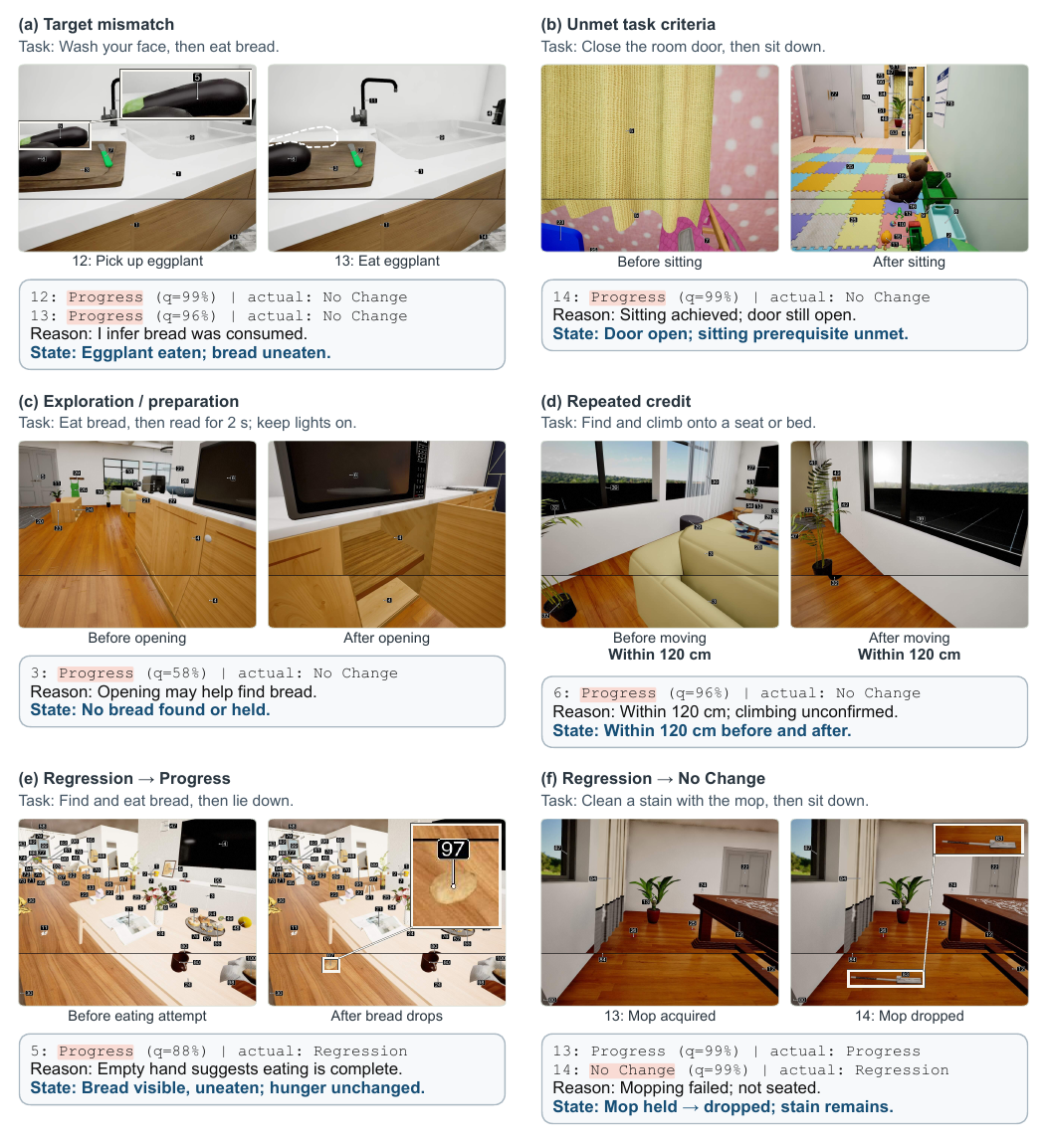}
  \caption{Examples of process-change misjudgments. Panels (a)--(d) illustrate four categories of false progress reports, and panels (e)--(f) illustrate two types of regression misjudgments. In (a), the white dashed line in the right image traces the eggplant's outline before pickup. Each row contrasts the reported category, confidence $q$, and actual change; pale red marks incorrect self-assessments.}
  \label{fig:cases}
\end{figure*}

\subsection{Overall Execution Performance and Full-Cycle Monitoring}

Execution performance and judgment accuracy are positively associated across models overall, with a stronger association for current-progress judgments than for online final-outcome forecasts (Fig.~\ref{fig:execution}). Within model families, Terra and Haiku outperform Sol and Sonnet, respectively, in task success rate, mean completion, and both judgment accuracy metrics, indicating that execution and self-assessment in embodied closed loops do not simply follow nominal model capability tiers. Despite differences in overall performance, the eight models exhibit consistent stage-dependent trends throughout execution: current-progress judgments become more prone to deviating from the actual state in later stages, whereas final-outcome judgments progressively approach actual outcomes but continue to overestimate final completion even at task termination (Fig.~\ref{fig:monitoring}).

For current-progress judgments, estimated completion generally rises as execution proceeds across all eight models, but overestimation becomes more pronounced in later stages than in earlier ones, with estimates exceeding actual completion by $10.6$--$22.0$ percentage points (Fig.~\ref{fig:monitoring}). This growing bias may be related to changes in the difficulty of verifying the task state as execution proceeds. Early on, most task requirements remain unsatisfied, so reporting no completed items often aligns with the actual state. As execution proceeds, satisfied and unsatisfied requirements coexist, requiring agents to verify both whether new actions have produced their intended effects and whether previously satisfied conditions still hold.

Final-outcome judgments, by contrast, exhibit the opposite trend. Before execution, models generally express optimistic expectations of near-complete success. As execution proceeds, their predicted final completion is gradually revised downward toward the actual outcome. Unlike current-progress judgments, which require verifying only the current state, final-outcome forecasts also depend on unexecuted actions and their uncertain outcomes. As execution unfolds, uncertainty decreases, allowing models to revise their initial expectations using more complete evidence of actual outcomes. At task termination, post-execution retrospective assessments are more accurate than pre-execution forecasts for all eight models, yet they still generally overestimate final completion.

Beyond this divergence in stage-dependent accuracy, calibration analysis reveals that confidence adjustments in both judgment types fail to track empirical accuracy across the eight models (Fig.~\ref{fig:monitoring}). For current-progress judgments, Gemini maintains confidence at $96.2\%$--$96.9\%$ throughout, while the other seven models reduce mean confidence in later stages, though by less than the corresponding drop in accuracy. In the late stage of execution, mean confidence exceeds empirical accuracy across all eight models; ECE also increases relative to the early stage in all models except Haiku. For final-outcome judgments, retrospective accuracy exceeds pre-execution forecast accuracy, but models also raise their confidence, leaving retrospective ECE at $9.2\%$--$35.4\%$. Overall, mean confidence generally exceeds empirical accuracy for both current-progress and final-outcome judgments across all eight models, indicating overconfidence.

\begin{figure*}[!t]
  \centering
  \includegraphics[width=\textwidth]{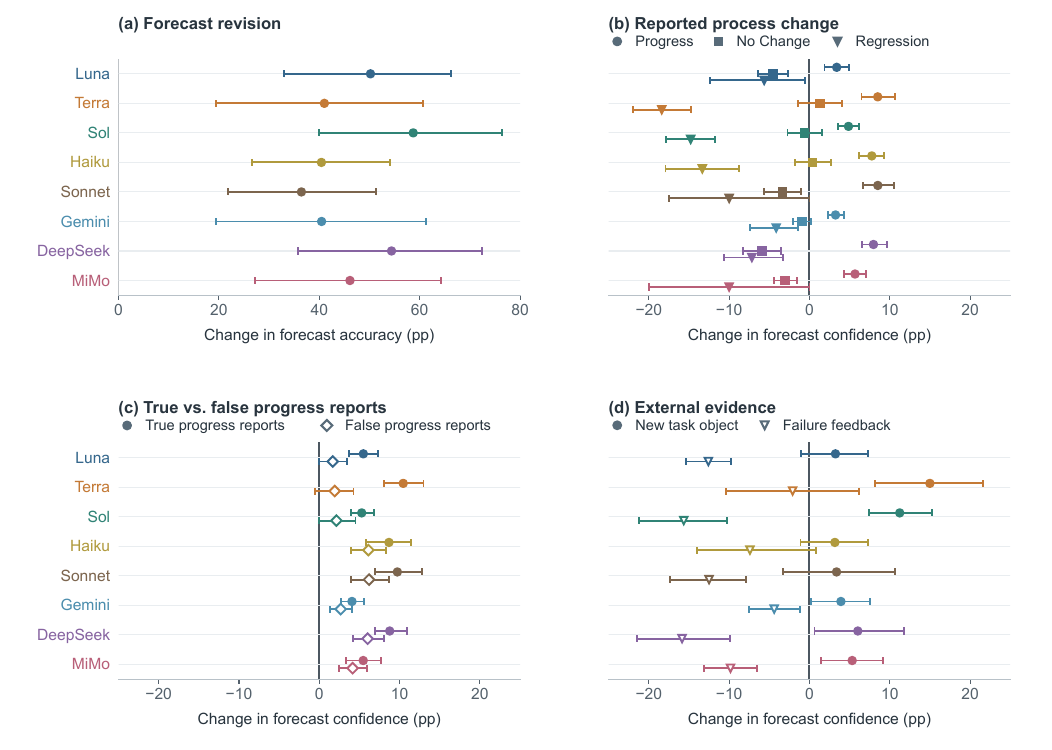}
  \caption{Final-outcome forecasts and confidence updates. (a) Change in forecast accuracy after revision relative to before revision. (b) Confidence updates associated with reported process changes. (c) Comparison of true and false progress reports. (d) Updates when new observations or failure feedback occur individually. Panel (a) uses adjacent reports with changed forecast content. Panels (b)--(d) show confidence differences between adjacent reports with unchanged forecast content. Error bars indicate 95\% confidence intervals.}
  \label{fig:confidence}
\end{figure*}

\subsection{Fine-Grained Change Recognition and Error Cases}

Analysis of process-change judgments shows that all eight models recognize actual progress relatively well, while errors concentrate in two patterns: misreporting \emph{No Change} as \emph{Progress}, and misclassifying \emph{Regression} as either \emph{No Change} or \emph{Progress} (Fig.~\ref{fig:process}a). Overall ECE for process-change judgments ranges from $5.6\%$ (Terra) to $30.3\%$ (DeepSeek).

For the first pattern, models not only frequently mistake unchanged states for progress but also substantially overestimate the reliability of their progress reports (Fig.~\ref{fig:process}b). Across all eight models, reported progress is less accurate than reported no change, yet mean confidence fails to decrease accordingly; by contrast, several models (notably Haiku, Sonnet, and MiMo) are underconfident when reporting no change. To identify the specific sources of false progress reports, we analyze records where actual \emph{No Change} is reported as \emph{Progress}, examining task requirements, pre- and post-action states, and reporting rationales. We defined the criteria for the following four attribution categories, and GPT-6.1-Sol applied them to classify each false progress report (Fig.~\ref{fig:errors}).

\begin{enumerate}

\item \textbf{Wrong target object.} Models count operations on non-target objects as task progress (Fig.~\ref{fig:cases}a). These errors account for $71.3\%$ and $66.7\%$ of Luna's and DeepSeek's false progress reports, respectively.

\item \textbf{Task conditions not met.} Models act on task-relevant objects but report progress even though required prerequisites remain unsatisfied or the action effect or interaction distance falls short of the prescribed criterion (Fig.~\ref{fig:cases}b). This category accounts for $53.6\%$ of Terra's false progress reports.

\item \textbf{Auxiliary or exploratory behavior counted as progress.} Models directly count preparatory or exploratory actions---such as turning, looking around, or opening cabinets to search---as progress, even though these actions establish no task-specified conditions (Fig.~\ref{fig:cases}c).

\item \textbf{Previously satisfied conditions counted again as progress.} Models count conditions that already held before an action as progress at the current step. For example, after moving, a model reports entering the required distance range even though it was already within that range prior to the action (Fig.~\ref{fig:cases}d).

\end{enumerate}

When established task conditions are lost, models likewise struggle to recognize regression, frequently misreporting it as \emph{Progress} or overlooking it as \emph{No Change} (Fig.~\ref{fig:process}a). For example, after a perturbation causes a piece of bread to be dropped, a model infers that eating is complete solely from the superficial cue of an empty hand and reports progress, despite an unchanged hunger level and the bread remaining clearly visible on the floor (Fig.~\ref{fig:cases}e). In another case, DeepSeek correctly reports progress after picking up a mop, but when a perturbation subsequently causes the mop to drop, the model focuses solely on the current action's failure and the remaining unsatisfied requirements; it consequently reports no change, overlooking the loss of the established possession condition, with confidence reaching $99\%$ in both judgments (Fig.~\ref{fig:cases}f).

\subsection{Final-Outcome Forecasts and Confidence Updates}

Comparing adjacent online reports along the same execution trajectory further reveals how models update both the content and confidence of their final-outcome forecasts. Forecast sets are revised in $12.5\%$ of adjacent report pairs. Among these revisions, mean final-outcome forecast accuracy improves by $36.4$--$58.7$ percentage points across the eight models (Fig.~\ref{fig:confidence}a), showing that models can correct some earlier forecast errors during execution. The remaining $87.5\%$ of report pairs retain the original forecast, and $90.9\%$ of these still predict that all task requirements will be satisfied. When forecast content remains unchanged, all eight models generally raise confidence after reporting \emph{Progress} and lower it after reporting \emph{Regression}, while confidence adjustments after reported \emph{No Change} vary across models (Fig.~\ref{fig:confidence}b).

When their final-outcome forecasts remain unchanged, all eight models show a mean increase in forecast confidence following both true and false progress reports, with a smaller increase following false reports (Fig.~\ref{fig:confidence}c). Moreover, when false progress reports coincide with forecast revisions, estimated final completion rises by an average of $30.6$ percentage points, whereas forecast accuracy drops by $27.3$ percentage points. Together, these results show that step-level misjudgments propagate to final-outcome forecasts, both inflating confidence in unchanged forecasts and misdirecting forecast revisions.

Beyond subjective process judgments, objective environmental evidence also drives confidence updates for existing final-outcome forecasts: all eight models generally raise confidence after first observing a task-relevant object and lower it after receiving an explicit execution error from the action interface (e.g., a target parameter incompatible with the requested action or an unreachable navigation destination) (Fig.~\ref{fig:confidence}d). Thus, while models can dynamically revise final-outcome expectations and confidence during execution using process judgments and environmental feedback, local misjudgments still disrupt this updating process. We next examine whether reported confidence can effectively distinguish correct from incorrect judgments.

\begin{table*}[!t]
  \centering
  \caption{AUROC of raw self-assessment confidence (\%, higher is better). Bold and underlined values denote the best and second-best pooled results in each column, respectively. Within denotes within-trajectory AUROC.}
  \label{tab:auroc}
  \small
  \setlength{\tabcolsep}{3pt}
  \begin{tabularx}{\textwidth}{@{}l*{8}{>{\centering\arraybackslash}X}@{}}
    \toprule
    Model & \shortstack{Pre-execution\\forecast} & \multicolumn{2}{c}{\shortstack{Online current-\\progress judgment}} & \multicolumn{2}{c}{\shortstack{Online process-\\change judgment}} & \multicolumn{2}{c}{\shortstack{Online final-\\outcome forecast}} & \shortstack{Retrospective\\assessment} \\
    \cmidrule(lr){3-4}\cmidrule(lr){5-6}\cmidrule(lr){7-8}
    & & Pooled & Within & Pooled & Within & Pooled & Within & \\
    \midrule
    Luna & \textbf{87.22} & \underline{88.89} & 79.75 & 74.71 & 69.24 & \textbf{81.48} & 82.78 & 80.55 \\
    Terra & 73.62 & \textbf{93.37} & 95.30 & \textbf{83.01} & 81.78 & 75.20 & 73.66 & \textbf{86.47} \\
    Sol & 77.23 & 88.72 & 85.46 & \underline{75.56} & 74.55 & \underline{79.50} & 78.53 & 81.68 \\
    Haiku & 75.79 & 87.38 & 81.13 & 60.95 & 53.93 & 72.79 & 74.06 & \underline{83.51} \\
    Sonnet & 79.74 & 85.94 & 82.23 & 64.13 & 59.17 & 71.97 & 77.96 & 81.22 \\
    Gemini & 71.54 & 63.28 & 56.25 & 53.49 & 49.65 & 72.53 & 69.90 & 68.92 \\
    DeepSeek & \underline{80.62} & 85.48 & 86.28 & 65.51 & 63.21 & 73.90 & 87.35 & 81.57 \\
    MiMo & 76.84 & 70.88 & 70.11 & 61.26 & 57.38 & 76.11 & 80.96 & 73.75 \\
    \bottomrule
  \end{tabularx}
\end{table*}

\subsection{Discrimination and Calibration of Self-Assessment Confidence}

Although self-assessment reports are generally overconfident, raw confidence may still provide a basis for selective verification and human intervention if it can rank the relative reliability of judgments. We therefore use AUROC to quantify confidence-based discrimination between correct and incorrect judgments at both the aggregate and individual-execution levels. \emph{Pooled AUROC} combines reports of the same type across trajectories to measure overall discrimination. \emph{Within-trajectory AUROC} is computed separately for each trajectory containing both correct and incorrect reports and then averaged, characterizing relative ranking ability within the same execution (Table~\ref{tab:auroc}).

Pooled AUROC exceeds the $50\%$ random-chance baseline across all eight models and self-assessment stages, ranging from $53.49\%$ to $93.37\%$, indicating that raw confidence carries useful signal regarding judgment correctness overall. Within-trajectory results further confirm that most models retain some relative ranking ability during individual executions. Across self-assessment dimensions, all models consistently show lower discrimination for online process-change judgments than for online current-progress judgments, suggesting that confidence-based ranking of single-step state transitions is more challenging. Gemini is particularly weak on both judgment types, with within-trajectory discrimination approaching chance.

\begin{table*}[!t]
  \centering
  \caption{ECE before and after one-dimensional calibration (\%, lower is better). Raw and Cal. denote original and calibrated confidence. Values are means over three task-level five-fold partitions.}
  \label{tab:one-dimensional-calibration}
  \small
  \setlength{\tabcolsep}{3pt}
  \begin{tabularx}{\textwidth}{@{}l*{10}{>{\centering\arraybackslash}X}@{}}
    \toprule
    Model & \multicolumn{2}{c}{\shortstack{Pre-execution\\forecast}} & \multicolumn{2}{c}{\shortstack{Online current-\\progress judgment}} & \multicolumn{2}{c}{\shortstack{Online process-\\change judgment}} & \multicolumn{2}{c}{\shortstack{Online final-\\outcome forecast}} & \multicolumn{2}{c}{\shortstack{Retrospective\\assessment}} \\
    \cmidrule(lr){2-3}\cmidrule(lr){4-5}\cmidrule(lr){6-7}\cmidrule(lr){8-9}\cmidrule(lr){10-11}
    & Raw & Cal. & Raw & Cal. & Raw & Cal. & Raw & Cal. & Raw & Cal. \\
    \midrule
    Luna & 35.28 & 7.64 & 16.20 & 7.37 & 27.86 & 7.02 & 34.80 & 5.57 & 27.03 & 9.18 \\
    Terra & 14.41 & 8.25 & 4.63 & 4.10 & 5.56 & 2.29 & 9.21 & 4.95 & 13.70 & 11.11 \\
    Sol & 34.47 & 8.45 & 6.96 & 4.44 & 9.12 & 4.18 & 27.05 & 4.74 & 18.96 & 12.03 \\
    Haiku & 26.10 & 12.93 & 8.78 & 6.54 & 9.18 & 3.77 & 15.38 & 5.62 & 9.22 & 9.17 \\
    Sonnet & 33.61 & 8.60 & 11.61 & 9.32 & 11.14 & 3.77 & 18.66 & 8.06 & 12.99 & 8.59 \\
    Gemini & 49.16 & 7.34 & 16.01 & 3.69 & 20.39 & 0.03 & 44.51 & 7.43 & 33.11 & 9.62 \\
    DeepSeek & 55.15 & 13.37 & 22.84 & 6.54 & 30.28 & 6.53 & 43.54 & 8.42 & 35.40 & 13.93 \\
    MiMo & 40.09 & 10.12 & 9.01 & 6.50 & 11.47 & 2.64 & 33.03 & 4.04 & 22.09 & 12.50 \\
    \bottomrule
  \end{tabularx}
\end{table*}

To examine whether post-hoc calibration can mitigate systematic bias in raw confidence, we fit a 1D monotonic mapping $\tilde q=\sigma(a\,\operatorname{logit}(q)+b)$ ($a\geq0$) for each model and self-assessment type while holding self-assessment content fixed. Evaluation uses task-grouped 5-fold cross-validation (with parameters estimated solely on training folds), averaging ECE across three fixed splits. As shown in Table~\ref{tab:one-dimensional-calibration}, ECE decreases consistently across all eight models and all stage-wise report types, providing a lightweight baseline for estimating judgment reliability from self-assessment signals. For Gemini's online process-change judgments, all calibrated confidence scores fall in a single bin, and their mean closely matches the observed accuracy, yielding an ECE of $0.03\%$.

\section{Discussion}
\label{sec:discussion}

\subsection{Stage-Dependent Biases and Cross-Granularity Error Propagation}

Current-progress judgments and final-outcome forecasts exhibit opposite accuracy trajectories across execution stages (Fig.~\ref{fig:monitoring}). Final-outcome forecasting benefits from progressive uncertainty resolution as remaining actions shrink and execution feedback accumulates, whereas current-progress judgment deteriorates under an increasing state-verification burden once satisfied and unsatisfied conditions coexist. Fine-grained error analysis shows that these failures stem from weak grounding of self-assessments in actual observations and task criteria---manifesting as target mismatches, unmet preconditions, and missed regressions of established states (Figs.~\ref{fig:process}--\ref{fig:cases}). Crucially, step-level misjudgments do not remain localized: when unchanged states are misreported as progress, the resulting errors propagate to task-level final-outcome forecasts, inflating predicted completion and driving unwarranted confidence increases (Fig.~\ref{fig:confidence}c).

\subsection{Dissociation Between Confidence Discrimination and Calibration}

Our evaluation highlights an important distinction between relative discrimination (AUROC) and absolute probability calibration (ECE). Although raw confidence systematically overestimates accuracy across all eight models, pooled and within-trajectory AUROC generally exceed chance (Table~\ref{tab:auroc}); moreover, during online updates, true progress reports induce larger confidence increases than false progress reports (Fig.~\ref{fig:confidence}c). Together, these static and dynamic patterns indicate that current models retain genuine, albeit imperfect, sensitivity to judgment validity despite severe miscalibration. Correspondingly, lightweight post-hoc monotonic calibration substantially lowers ECE across all models (Table~\ref{tab:one-dimensional-calibration}), although it addresses only systematic scaling bias rather than underlying discrimination.

\subsection{Implications for Embodied Self-Monitoring and Human--Robot Collaboration}

These findings carry two concrete implications for embodied agent design and human--robot teaming. First, because late-stage overestimation and cross-granularity error propagation stem from unverified step-level claims, embodied self-monitoring should enforce condition-level verification against current observations and maintain explicit checks on previously satisfied conditions. Second, in human--robot collaboration, communicating raw verbalized confidence directly to operators risks inducing miscalibrated reliance and over-trust. Given that most models preserve non-trivial within-trajectory ranking signal, a more principled near-term practice is to use post-hoc calibrated confidence as a relative risk indicator for triggering selective re-verification or event-triggered human intervention~\cite{conlon2024}.

\subsection{Limitations}

First, our evaluation is conducted on TongSIM, where discrete action interfaces and controlled state perturbations cannot fully reproduce the continuous-control dynamics and sensor noise of real-world robotic deployment. Second, agents receive only a single first-person RGB frame per round alongside text-based episodic history; how full-cycle self-monitoring behaves under continuous video streams or multimodal episodic memory remains an open question for future work.
\section{Conclusion}
\label{sec:conclusion}

We introduced EmbodiedMonitor, a benchmark for embodied self-monitoring throughout task execution that jointly evaluates judgment accuracy and confidence calibration across stages along the same trajectory. Closed-loop evaluation of eight widely used multimodal models shows that, although current models exhibit partial self-assessment and confidence-discrimination capabilities, late-stage progress overestimation, false progress reports, missed regressions, and systematic overconfidence prevent them from directly supporting reliable trust calibration in human--robot collaboration. EmbodiedMonitor provides a standardized empirical foundation for advancing observation-grounded verification and calibrated self-monitoring in embodied agents.


\begin{thebibliography}{99}

\bibitem{norton2022}
A. Norton \emph{et al.}, ``Metrics for robot proficiency self-assessment and communication of proficiency in human-robot teams,'' \emph{ACM Trans. Human-Robot Interact.}, vol. 11, no. 3, Art. no. 29, 2022.

\bibitem{conlon2022}
N. Conlon, D. Szafir, and N. Ahmed, ```I'm confident this will end poorly': Robot proficiency self-assessment in human-robot teaming,'' in \emph{Proc. IEEE/RSJ Int. Conf. Intell. Robots Syst. (IROS)}, 2022, pp. 2127--2134.

\bibitem{conlon2024}
N. Conlon, N. Ahmed, and D. Szafir, ``Event-triggered robot self-assessment to aid in autonomy adjustment,'' \emph{Frontiers Robot. AI}, vol. 10, Art. no. 1294533, Jan. 2024, doi: 10.3389/frobt.2023.1294533.

\bibitem{frasca2020}
T. Frasca \emph{et al.}, ```Can you do this?' Self-assessment dialogues with autonomous robots before, during, and after a mission,'' in \emph{HRI Workshop on Assessing, Explaining, and Conveying Robot Proficiency for Human-Robot Teaming}, 2020, arXiv:2005.01544.

\bibitem{embodiedbench2025}
R. Yang \emph{et al.}, ``EmbodiedBench: Comprehensive benchmarking multi-modal large language models for vision-driven embodied agents,'' in \emph{Proc. Int. Conf. Mach. Learn. (ICML)}, 2025.

\bibitem{eai2024}
M. Li \emph{et al.}, ``Embodied agent interface: Benchmarking LLMs for embodied decision making,'' in \emph{Advances in Neural Information Processing Systems, Datasets and Benchmarks Track}, 2024.

\bibitem{yu2025}
T. Yu \emph{et al.}, ``Uncertainty in action: Confidence elicitation in embodied agents,'' arXiv:2503.10628, 2025.

\bibitem{wei2026scoper}
Y. Wei, Y. Fang, S. Yang, T. Wu, T. Wang, and X. Qi, ``Towards a systematic study of embodied system 2 over full task execution cycles,'' arXiv:2610.17657, 2026.

\bibitem{tongsim2025}
Z. Sun \emph{et al.}, ``TongSIM: A general platform for simulating intelligent machines,'' arXiv:2512.20206, 2025.

\bibitem{shridhar2020}
M. Shridhar \emph{et al.}, ``ALFRED: A benchmark for interpreting grounded instructions for everyday tasks,'' in \emph{Proc. IEEE/CVF Conf. Comput. Vis. Pattern Recognit. (CVPR)}, 2020, pp. 10740--10749.

\bibitem{shridhar2021}
M. Shridhar \emph{et al.}, ``ALFWorld: Aligning text and embodied environments for interactive learning,'' in \emph{Proc. Int. Conf. Learn. Represent. (ICLR)}, 2021.

\bibitem{padmakumar2022}
A. Padmakumar \emph{et al.}, ``TEACh: Task-driven embodied agents that chat,'' in \emph{Proc. AAAI Conf. Artif. Intell.}, vol. 36, no. 2, 2022, pp. 2017--2025.

\bibitem{srivastava2022}
S. Srivastava \emph{et al.}, ``BEHAVIOR: Benchmark for everyday household activities in virtual, interactive, and ecological environments,'' in \emph{Proc. Conf. Robot Learn. (CoRL)}, \emph{Proc. Mach. Learn. Res.}, vol. 164, 2022, pp. 477--490.

\bibitem{li2023}
C. Li \emph{et al.}, ``BEHAVIOR-1K: A benchmark for embodied AI with 1,000 everyday activities and realistic simulation,'' in \emph{Proc. Conf. Robot Learn. (CoRL)}, \emph{Proc. Mach. Learn. Res.}, vol. 205, 2023, pp. 80--93.

\bibitem{szot2021}
A. Szot \emph{et al.}, ``Habitat 2.0: Training home assistants to rearrange their habitat,'' in \emph{Advances in Neural Information Processing Systems}, vol. 34, 2021, pp. 251--266.

\bibitem{steyvers2026}
M. Steyvers and M. A. K. Peters, ``Metacognition and uncertainty communication in humans and large language models,'' \emph{Current Directions Psychol. Sci.}, vol. 35, no. 3, 2026, doi: 10.1177/09637214251391158.

\bibitem{guo2017}
C. Guo, G. Pleiss, Y. Sun, and K. Q. Weinberger, ``On calibration of modern neural networks,'' in \emph{Proc. Int. Conf. Mach. Learn. (ICML)}, \emph{Proc. Mach. Learn. Res.}, vol. 70, 2017, pp. 1321--1330.

\bibitem{kadavath2022}
S. Kadavath \emph{et al.}, ``Language models (mostly) know what they know,'' arXiv:2207.05221, 2022.

\bibitem{lin2022}
S. Lin, J. Hilton, and O. Evans, ``Teaching models to express their uncertainty in words,'' \emph{Trans. Mach. Learn. Res.}, 2022.

\bibitem{tian2023}
K. Tian \emph{et al.}, ``Just ask for calibration: Strategies for eliciting calibrated confidence scores from language models fine-tuned with human feedback,'' in \emph{Proc. Conf. Empirical Methods Natural Lang. Process. (EMNLP)}, 2023, pp. 5433--5442.

\bibitem{xiong2024}
M. Xiong \emph{et al.}, ``Can LLMs express their uncertainty? An empirical evaluation of confidence elicitation in LLMs,'' in \emph{Proc. Int. Conf. Learn. Represent. (ICLR)}, 2024.

\bibitem{ding2025}
Z. Ding, S. Ni, and K. Bi, ``Do LVLMs know what they know? A systematic study of knowledge boundary perception in LVLMs,'' in \emph{Findings Assoc. Comput. Linguist.: EMNLP}, 2025, pp. 19832--19848.

\bibitem{israelsen2025famsec}
B. Israelsen \emph{et al.}, ```A good bot always knows its limitations': Assessing autonomous system decision-making competencies through factorized machine self-confidence,'' \emph{ACM Trans. Human-Robot Interact.}, vol. 14, no. 4, pp. 1--63, 2025, doi: 10.1145/3732794.

\bibitem{warmsley2025}
D. Warmsley, K. Choudhary, J. Rego, E. Viani, and P. K. Pilly, ``Self-assessment in machines boosts human trust,'' \emph{Frontiers Robot. AI}, vol. 12, Art. no. 1557075, 2025.

\bibitem{singh2026brace}
J. Singh \emph{et al.}, ``Agent-BRACE: Decoupling beliefs from actions in long-horizon tasks via verbalized state uncertainty,'' arXiv:2605.11436, 2026.

\bibitem{hathidara2026}
A. Hathidara, ``Can an agent learn to predict its own failure? Outcome-grounded confidence calibration in LLM agents,'' arXiv:2610.18614, 2026.

\bibitem{cote2018textworld}
M.-A. C\^ot\'e \emph{et al.}, ``TextWorld: A learning environment for text-based games,'' in \emph{IJCAI Computer Games Workshop}, 2018, arXiv:1806.11532.

\bibitem{yao2022webshop}
S. Yao, H. Chen, J. Yang, and K. Narasimhan, ``WebShop: Towards scalable real-world web interaction with grounded language agents,'' in \emph{Advances in Neural Information Processing Systems}, vol. 35, 2022.

\bibitem{johnson2016malmo}
M. Johnson, K. Hofmann, T. Hutton, and D. Bignell, ``The Malmo platform for artificial intelligence experimentation,'' in \emph{Proc. Int. Joint Conf. Artif. Intell. (IJCAI)}, 2016.

\bibitem{deitke2022procthor}
M. Deitke \emph{et al.}, ``ProcTHOR: Large-scale embodied AI using procedural generation,'' in \emph{Advances in Neural Information Processing Systems}, vol. 35, 2022.

\end{thebibliography}
\end{document}